\documentclass{article}
\usepackage{spconf,amsmath,amssymb,graphicx,booktabs,multirow,xcolor}
\usepackage{arydshln}

\usepackage[hidelinks]{hyperref}
\usepackage{microtype}
\usepackage{tikz}
\usetikzlibrary{arrows.meta,positioning,fit,calc,backgrounds}

\newcommand{\bench}{JusticeAxis}
\newcommand{\benchshort}{JusticeAxis}
\newcommand{\method}{JusticeAgent}
\newcommand{\task}{Reference-Anchored Legal Judgment}

\newcommand{\X}{\mathbb{X}}

\DeclareMathOperator*{\argmax}{arg\,max}

\newcommand{\Jgold}{J^{*}}
\newcommand{\Jrig}{J^{\mathrm{rig}}}
\newcommand{\Jung}{J^{\mathrm{ung}}}
\newcommand{\gv}{\,|\,}      
\newcommand{\LCB}{\mathrm{LCB}}
\newcommand{\UCB}{\mathrm{UCB}}

\title{JusticeAxis: Benchmarking Legal Judgment between Rigid Rule Application and Ungrounded Discretion}

\name{Zhengkai Tu$^{1}$, Mingda Zhang$^{2}$, Zijia Wang$^{3}$, Xiaoying Tang$^{2}$, Jimmy Huang$^{4,*}$\thanks{$^{*}$Corresponding author: \texttt{jhuang@yorku.ca}.}}
\address{$^{1}$The Chinese University of Hong Kong \quad $^{2}$The Chinese University of Hong Kong, Shenzhen\\
  $^{3}$University of Oxford \quad $^{4}$York University\\[3pt]
  \texttt{mingdazhang@ieee.org}}

\begin{document}

\ninept
\maketitle

\begin{abstract}
A sound judgment applies the law to established facts and weighs the circumstances in which they arose.
However, existing methods swing between rigid statute matching and ungrounded discretion, benchmarks score a label or a rubric, and the experience that would supply the balance stays unverified.
We formalize legal judgment as a reference-anchored task, whose object is a single decision that stays tied to the statute and to the circumstances at once.
We introduce \bench{}, 256 real-world criminal cases from 18 countries with audio, image, and text evidence, and three lawyer-written judgments for every case: the recorded one and one for each failure.
We further propose \method{}, a harness whose element agents establish the facts and whose judge agent applies the law under skills carrying experience of the circumstances.
Skills are distilled from execution trajectories and admitted only under Bayesian credible bounds.
Experiments show that failure turns direction with scale: open-weight backbones drift to unsupported grounds, frontier models to the statutory default.
We further verify that \method{}, as a simple yet effective plugin, carries a frozen open-weight backbone to commercial level.
Project resources are available at \url{https://github.com/beita6969/JusticeAxis}.
\end{abstract}


\section{Introduction}
\label{sec:intro}

\begin{figure}[t]
  \centering
  \includegraphics[width=\linewidth]{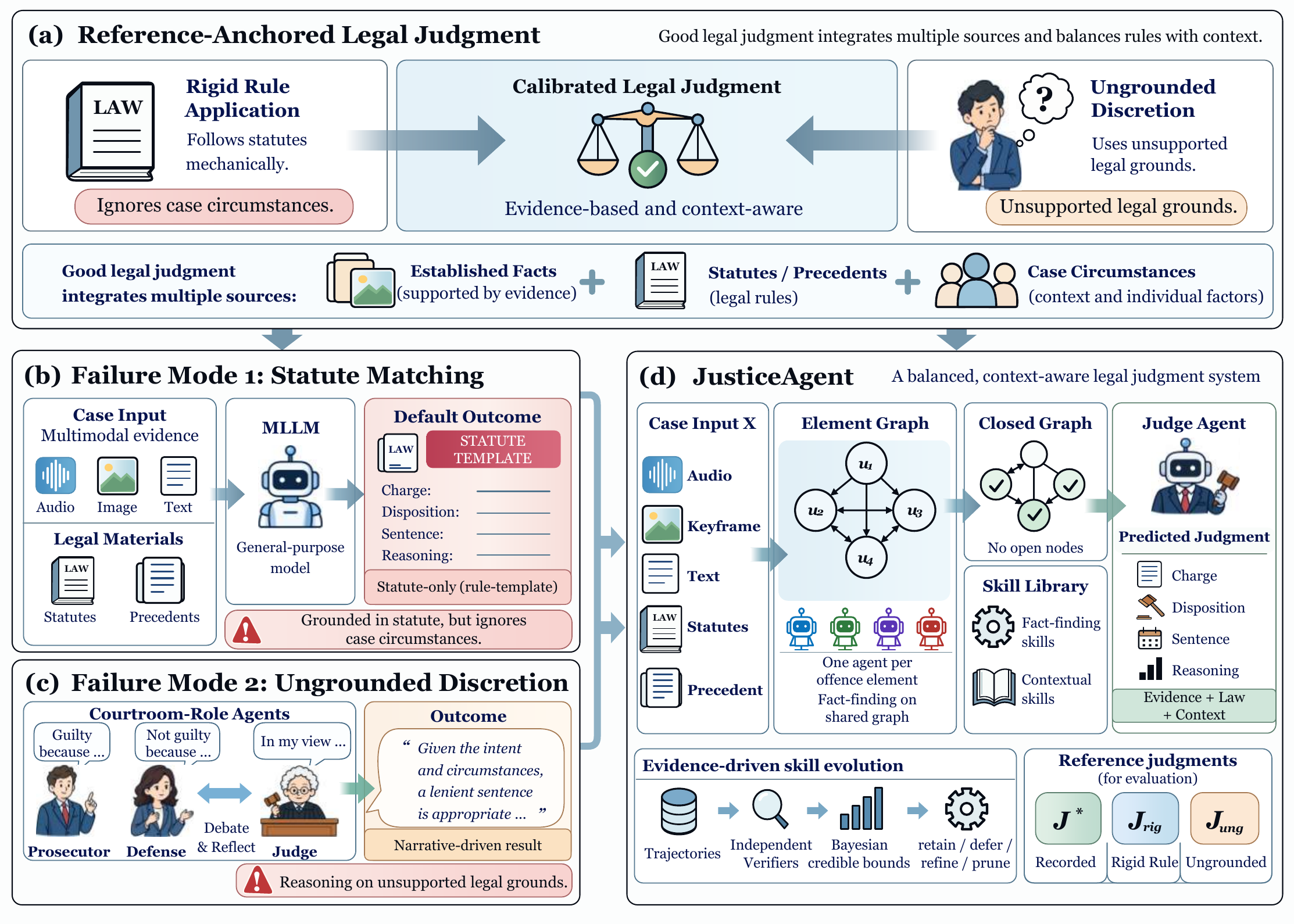}
  \caption{Overview: the axis a judgment sits on (a), the two failure modes of existing systems, rigid statute matching (b) and ungrounded discretion (c), and the \method{} framework (d).}
  \label{fig:overview}
\end{figure}

In recent years, LLM-based agents have been applied to legal judgment prediction, which predicts the charge, the statutory provisions, and the sentence of a case~\cite{xiao2018cail2018,fei2024lawbench,kang2026multimodal,liao2026verdict}, a key entry point from document processing to decision support.
Such a judgment requires facts established from the evidence and checked against the statutes, and then a choice between a rule whose text is fixed before the case and a standard whose content is settled only against it~\cite{kaplow1992rules,casey2016death,shen2025law}.
An agent reading the record of a case to determine the charge and the outcome it carries could support courts, and defendants with little access to legal help (Fig.~\ref{fig:overview}a)~\cite{gonzalez2026egopolice}.
Unlike tasks with a single correct label, the same charge leads to markedly different dispositions and sentences depending on culpability, harm, and local practice~\cite{sentencingcouncil2019}, posing a core challenge for agent-driven adjudication.

Surprisingly, however, we observed that the two failures divide by scale: open-weight backbones argue from unsupported grounds (Fig.~\ref{fig:overview}c), while commercial models return the statute's default outcome where they miss (Fig.~\ref{fig:overview}b).
Their judgments then diverge from what a court would decide, with the consequences borne by defendants~\cite{linna2026challenges}.
Meanwhile, to the best of our knowledge, how far a system sits between rigid rule application and ungrounded discretion remains to be quantitatively compared across models~\cite{xiao2018cail2018,fei2024lawbench,ovcharov2026multi}.
Furthermore, courtroom-role pipelines (Fig.~\ref{fig:overview}c) also warrant comparison against element-based approaches (Fig.~\ref{fig:overview}d)~\cite{kang2026multimodal,liao2026verdict,he2024agentscourt}.

Therefore, a critical research question raises:
\emph{How can models be guided to apply the law to established facts while weighing the circumstances, and how can this balance be quantitatively assessed?}

To address this issue, we introduce \benchshort{}, a benchmark centred on courts establishing the facts of a case from its evidence and then applying the law under the experience the circumstances call for, and we perform a series of confirmatory experiments (Fig.~\ref{fig:overview}).
The main contributions are as follows:
\begin{itemize}\setlength{\itemsep}{0pt}\setlength{\parskip}{0pt}
\item \textbf{\task{} Task:} We formalize legal judgment as a single pass producing the charge, the disposition, the sentence, and the reasoning, read against three lawyer-written references for the same case: rigid rule application, ungrounded discretion, and the judgment the court recorded.
\item \textbf{\bench{} Benchmark:} We construct \bench{}, the first benchmark to supply, for every case, a written judgment for each failure mode, comprising 256 real-world criminal cases from 18 countries with lawyer-written reference judgments.
\item \textbf{\method{} Framework:} We propose \method{}, a multi-agent harness in which one agent per element of the offence establishes the facts on a shared graph and a judge agent applies the law under verified experience, lifting a frozen open-weight backbone to commercial level in a plug-in way.
\end{itemize}

\section{Related Work}
\label{sec:related}

\noindent\textbf{Agents for legal judgment.}
Early systems prompt one model over retrieved statutes and precedents~\cite{jiang2023legal,zhang2025syler,yang2026glare}; later ones organise several agents as a courtroom to debate and deliberate~\cite{he2024agentscourt,chen2025agentcourt,zhang2026chinese,liao2026verdict,kang2026multimodal}, assign one agent per rule element~\cite{yuan2024can}, or let a skill library evolve across cases~\cite{geng2026parthenon,wu2026bayesian}; \method{} builds on both.

\noindent\textbf{Benchmarks for legal reasoning.}
Benchmarks have moved from charge and article classification~\cite{xiao2018cail2018,fei2024lawbench} to broad task suites~\cite{guha2023legalbench,li2025legalagentbench}, exam-style reasoning~\cite{fan2026lexam,pires2026magis}, cross-jurisdictional text~\cite{ovcharov2026multi,yang2026crosslex}, and citation grounding~\cite{ovcharov2026citation,choudhury2026better}.
Released sets are almost all textual~\cite{hou2025large}, and courtroom-speech corpora serve conversation analysis and outcome prediction, not judgment~\cite{danescu2012echoes}.
All score a produced label or a rubric~\cite{liu2026llm}; none supplies, per case, a written judgment for each way a system can fail, which makes the balance measurable.

\section{Task and Benchmark}
\label{sec:task}

\begin{figure*}[!t]
  \centering
  \includegraphics[width=\textwidth]{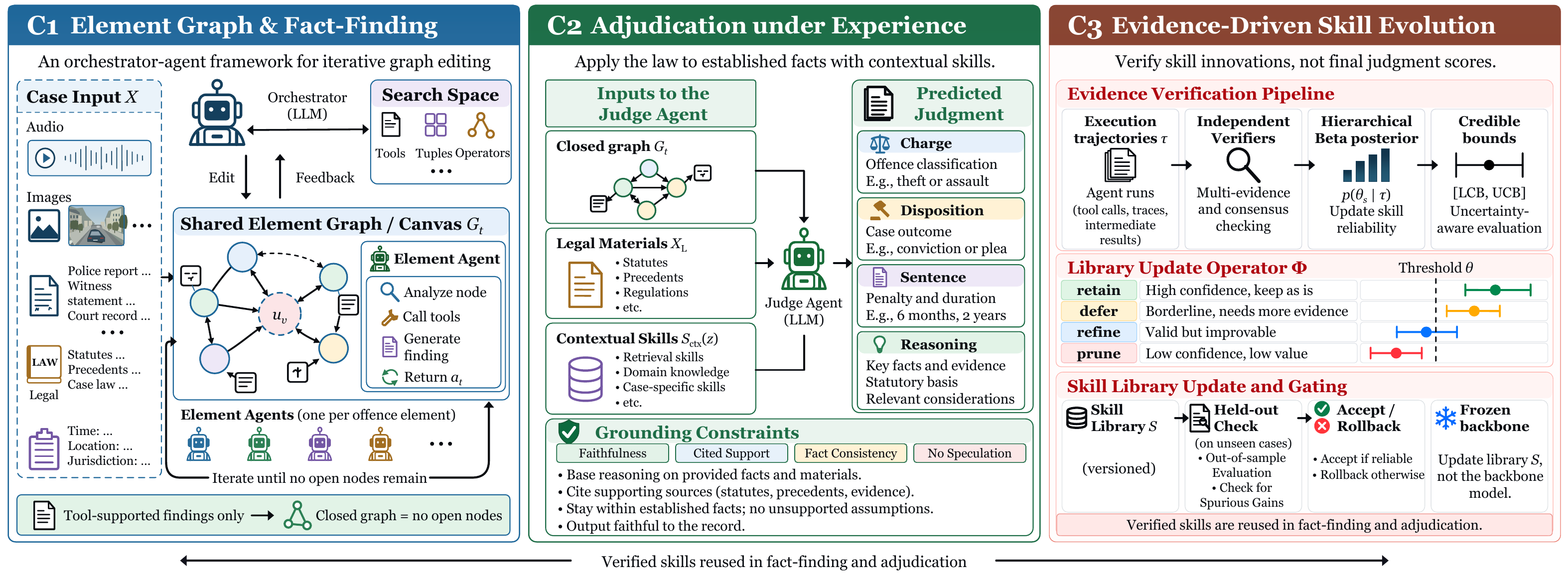}
  \caption{The \method{} harness. \textbf{C1}: an orchestrator edits a shared Element Graph while one agent per offence element calls tools and returns findings until no node is open. \textbf{C2}: the judge applies the law to the closed graph under contextual skills and grounding constraints. \textbf{C3}: skills mined from trajectories are verified independently and retained, deferred, refined or pruned on credible bounds, not on judgment scores.}
  \label{fig:method}
\end{figure*}

\noindent\textbf{Reference-anchored legal judgment.}
As in Fig.~\ref{fig:overview}a, we define one task along a court's path, from evidence to a judgment checkable on grounding and fit to the circumstances, with a single input:
\begin{equation}
  \X = \{X_m \gv m \in \mathcal{M}\},\qquad \mathcal{M}\subseteq\{A, I, T, L\},
  \label{eq:input}
\end{equation}
where $X_A$, $X_I$ and $X_T$ are the audio, the keyframes, and the text evidence together with the background of the case, and $X_L$ the candidate statutes and the comparable precedents the charge turns on; missing modalities are allowed.
Given $\X$, a system generates the judgment $J=(c,d,s,r)$, namely the charge, the disposition, the sentence, and the reasoning that cites what it relies on:
\begin{equation}
  \hat{J} \;=\; \argmax_{J}\ \Pr(J \gv \X).
  \label{eq:task}
\end{equation}
The recorded charge is never supplied, so a system must establish what happened before deciding which statute governs it; a statute governs only if all its elements are established.

\noindent\textbf{Anchored objective.}
Each case carries three judgments written by law professors and practising lawyers: $\Jgold$ as the court decided it, $\Jrig$ applying the matched statute to the established facts regardless of circumstances, and $\Jung$ arguing from the narrative on unsupported legal bases.
A judgment is placed by the reference it lands closest to under the outcome distance $\rho$ below; the shares across the three references locate a system on the axis of Fig.~\ref{fig:overview}a, and the signed share of the misses
\begin{equation}
  \mathrm{Pol}=\frac{|\{a=\Jrig\}|-|\{a=\Jung\}|}{|\{a\neq\Jgold\}|}\in[-1,1]
  \label{eq:pol}
\end{equation}
says which way it fails, $+1$ reciting the statute and $-1$ arguing from unsupported grounds.
Two poles rather than one follow a standard account of discretion as an area left open by a belt of restriction~\cite{dworkin1977taking}: leaving the belt and standing still inside it are different errors.

\noindent\textbf{The \bench{} benchmark.}
We construct \bench{} from 256 concluded criminal cases whose recordings or footage were publicly released, spanning 18 countries and 18 charge types (Table~\ref{tab:stats}); charges recur with different circumstances and recorded outcomes.
Public releases carry the recording alone, so each case is completed into one document: audio and keyframe references with their written descriptions, the background of the case, and the legal materials $X_L$; reconstructed background is marked with its source and the limits of the inference; participants are aliased and the recorded outcome is withheld from the solver input.
Every annotation is written by hand under a two-stage protocol by law professors and practising lawyers: source-attributed evidence statements decomposed into atomic fact units, then the three reference judgments with their reasoning, each failure judgment carrying the step at which it goes wrong~\cite{choudhury2026better}.
A second annotator reviews every case; answers are sealed during inference.

\begin{table}[t]
  \centering\footnotesize
  \setlength{\tabcolsep}{2pt}
  \renewcommand{\arraystretch}{0.95}
  \caption{Comparison with legal and multimodal benchmarks. A/I/T: audio, image, text; Law, Prec., Bg.: statutes, precedents, background; Refs: references per case; Fail.: who wrote the failure ones.}
  \label{tab:stats}
  \begin{tabular}{@{}l c c c c c c l@{}}
    \toprule
    & & & \multicolumn{3}{c}{Given per instance} & & \\
    \cmidrule(lr){4-6}
    Benchmark & Jur. & Input & Law & Prec. & Bg. & Refs & Fail. \\
    \midrule
    CAIL2018~\cite{xiao2018cail2018}      & 1  & T     &            &            &            & 1      & \\
    LegalBench~\cite{guha2023legalbench}  & 1  & T     &            &            &            & 1      & \\
    CrossLex~\cite{yang2026crosslex}      & 3  & T     & \checkmark &            &            & 1      & \\
    Multi-Legal-Bench~\cite{ovcharov2026multi}  & 6  & T     &            &            &            & 1      & \\
    CLAUSE~\cite{choudhury2026better}          & 1  & T     & \checkmark &            &            & 1      & LLM \\
    Magis-Bench~\cite{pires2026magis} & 1  & T     &            &            &            & rubric & \\
    JurisMM~\cite{kang2026multimodal}       & 1  & I+T   &            &            &            & 1      & \\
    EgoPolice~\cite{gonzalez2026egopolice}    & 1  & A+I   &            &            &            & 1      & \\
    \midrule
    \bench{} (ours) & 18 & A+I+T & \checkmark & \checkmark & \checkmark & 3 & lawyers \\
    \bottomrule
  \end{tabular}
  \vspace{-2mm}
\end{table}

\noindent\textbf{Metrics.}
Charges are scored by exact-match and family-level accuracy and outcomes by disposition and exact-sentence accuracy~\cite{fei2024lawbench}, each also conditioned on the coarser decision being right.
As in CAIL2018~\cite{zhong2018overview}, sentences are compared by the log-difference $\ell(s,s')=|\log(1+s)-\log(1+s')|$, and
\begin{equation}
  \rho(J,J')=\lambda_d\mathbf{1}[d\neq d']+\lambda_s\,\ell(s,s')/\log(1+s_{\max})
  \label{eq:rho}
\end{equation}
carries it with disposition mismatch into the outcome distance of Eq.~\eqref{eq:pol}, $s_{\max}$ being the longest sentence in the data.
Anchoring compares outcomes; how a judgment is argued is constrained rather than scored (Sec.~\ref{ssec:judge})~\cite{ovcharov2026citation}, so fluency earns no credit.

\section{JusticeAgent}
\label{sec:method}

Reading the record once establishes no facts and weighs no circumstances, the two failures of Sec.~\ref{sec:task}.
\method{} is a harness~\cite{guo2026question} around a frozen backbone $\mathcal{M}_{\mathrm{exec}}$ separating them (Fig.~\ref{fig:method}).

\subsection{Element Graph and Fact-Finding}
\label{ssec:graph}

\noindent\textbf{Definition 1 (Element Graph).}
An Element Graph is a directed acyclic graph $\mathcal{G}=(\mathcal{V},\mathcal{E},\mathrm{attr})$ whose nodes are the elements of the offence required by $X_L$, whose edges encode legal dependency (injury $\to$ seriousness $\to$ charge), and whose attributes
\begin{equation}
  \mathrm{attr}(v)=\bigl(u_v,\ \mathcal{F}_v,\ f_v,\ \sigma_v\bigr),
  \label{eq:node}
\end{equation}
record the element $u_v$, the evidence $\mathcal{F}_v$ gathered for it, the finding $f_v$ with its cited evidence chain, and a status $\sigma_v$ that is open, found, or unestablished; a graph is \emph{closed} when no node is open.

\noindent\textbf{Environment.}
The harness holds the graph and callable resources:
\begin{equation}
  \mathcal{H}=\bigl(\mathcal{G}_t,\ \mathcal{S},\ \mathcal{T},\ \mathcal{V}_{\mathrm{ver}},\ \mathcal{M}_{\mathrm{exec}}\bigr),
  \label{eq:harness}
\end{equation}
where $\mathcal{G}_t$ is the graph after $t$ edits, $\mathcal{S}$ the skill library, $\mathcal{T}$ the transcription, keyframe, retrieval and alignment tools, and $\mathcal{V}_{\mathrm{ver}}$ the verifiers.

\noindent\textbf{Orchestrator and Element Agents.}
From $\mathcal{G}_0$, built from $X_L$, the orchestrator commits one atomic edit $a_t$ per turn, of type $\alpha_t$, and a dispatched node $v$ goes to an element agent that issues at most $K$ tool calls $b_{v,1:K}$ and returns a finding with its evidence chain:
\begin{equation}
  \mathcal{G}_t=\mathcal{G}_{t-1}\oplus a_t,\quad
  (f_v,\sigma_v)\sim\pi_{\mathrm{elem}}\bigl(\cdot\gv u_v,\mathcal{F}_v(b_{v,1:K}),\mathcal{S}_{u_v}\bigr),
  \label{eq:update}
\end{equation}
where $\mathcal{S}_{u_v}\subseteq\mathcal{S}$ are the skills retrieved for $u_v$; an agent may assert only what a tool output or attributed fact supports, and the orchestrator edits until the graph is closed.
A trajectory $\tau=\{(a_t,o_t)\}_{t=1}^{T}$ ends with $\alpha_T=\mathrm{adjudicate}$ and factorises as
\begin{equation}
  P(\tau\gv\X)=\!\prod_{t=1}^{T}\!
  \pi_{\mathrm{orch}}(a_t\gv\mathcal{G}_{t-1},o_{<t})\!
  \pi_{\mathrm{elem}}(o_t\gv a_t,\mathcal{G}_{t-1},\X,\mathcal{S}).
  \label{eq:traj}
\end{equation}

\subsection{Adjudication under Experience}
\label{ssec:judge}

\begin{table*}[t]
  \centering\scriptsize
  \renewcommand{\arraystretch}{1.18}
  \caption{Main results (128 evaluation cases). Acc, Fam, Disp and Sent.\ are the shares of cases with charge, charge family, disposition and sentence exactly correct, Avg their mean; Acc/Fam is exact charge among family-correct cases, Sent/Disp exact sentence among disposition-correct cases. Nearest anchor is the share landing closest to each reference under $\rho$, Pol.\ the signed share of misses toward the rigid pole, $(\Jrig-\Jung)/(\Jrig+\Jung)$. Miss is $100-\Jgold$. $\pm$ are Wilson 95\% half-widths; $\Delta$ is the gain over the Qwen3.8-27B backbone, RMR the relative reduction of its misses. Higher is better except $\Jrig$, $\Jung$, Miss. $^\dagger$304B mixture-of-experts.}
  \label{tab:main}
  \setlength{\tabcolsep}{2.1pt}
  \begin{tabular*}{\textwidth}{@{\extracolsep{\fill}} l ccccc cc cccc c cccc}
    \toprule
    & \multicolumn{5}{c}{Correct (\%)} & \multicolumn{2}{c}{Conditional (\%)} & \multicolumn{4}{c}{Nearest anchor (\%)} & & \multicolumn{4}{c}{$\Delta$ vs backbone}\\
    \cmidrule(lr){2-6}\cmidrule(lr){7-8}\cmidrule(lr){9-12}\cmidrule(lr){14-17}
    Model & Acc & Fam & Disp & Sent. & Avg & Acc/Fam & Sent/Disp & $\Jgold$ & $\Jrig\downarrow$ & $\Jung\downarrow$ & Miss$\downarrow$ & Pol. & Acc & Sent. & $\Jgold$ & RMR \\
    \midrule
    \multicolumn{17}{l}{\emph{Open-weight MLLMs, 27--38B}}\\
    Qwen3.8-27B & 47.66{\tiny$\pm$8.53} & 56.25 & 62.50 & 28.91 & 48.83 & 84.72 & 46.25 & 54.69{\tiny$\pm$8.50} & 14.84 & 30.47 & 45.31 & -0.345{\tiny$\pm$0.235} & -- & -- & -- & -- \\
    Gemma 4 31B & 32.81{\tiny$\pm$8.03} & 37.50 & 39.84 & 17.97 & 32.03 & 87.50 & 45.10 & 34.38{\tiny$\pm$8.12} & 20.31 & 45.31 & 65.62 & -0.381{\tiny$\pm$0.194} & -- & -- & -- & -- \\
    DeepSeek-V4-Flash$^\dagger$ & 59.38{\tiny$\pm$8.39} & 64.06 & 73.44 & 36.72 & 58.40 & 92.68 & 50.00 & 66.41{\tiny$\pm$8.08} & 11.72 & 21.88 & 33.59 & -0.302{\tiny$\pm$0.274} & -- & -- & -- & -- \\
    \hdashline[1pt/1.5pt]
    \multicolumn{17}{l}{\emph{Commercial MLLMs}}\\
    Claude Sonnet 5 & 78.12{\tiny$\pm$7.10} & 81.25 & 86.72 & 67.97 & 78.52 & 96.15 & 78.38 & 84.38{\tiny$\pm$6.28} & 11.72 & 3.91 & 15.62 & 0.500{\tiny$\pm$0.357} & -- & -- & -- & -- \\
    Gemini 3.8 Flash & 82.81{\tiny$\pm$6.51} & 88.28 & 92.19 & 74.22 & 84.38 & 93.81 & 80.51 & 90.62{\tiny$\pm$5.11} & 5.47 & 3.91 & 9.38 & 0.167{\tiny$\pm$0.487} & -- & -- & -- & -- \\
    Grok 4.6 & 79.69{\tiny$\pm$6.92} & 85.94 & 91.41 & 75.78 & 83.20 & 92.73 & 82.91 & 89.84{\tiny$\pm$5.29} & 7.81 & 2.34 & 10.16 & 0.538{\tiny$\pm$0.421} & -- & -- & -- & -- \\
    GPT-5.6 & 89.06{\tiny$\pm$5.45} & 92.97 & 96.88 & 85.16 & 91.02 & 95.80 & 87.90 & 93.75{\tiny$\pm$4.32} & 5.47 & 0.78 & 6.25 & 0.750{\tiny$\pm$0.448} & -- & -- & -- & -- \\
    \hdashline[1pt/1.5pt]
    \multicolumn{17}{l}{\emph{Peer frameworks on Qwen3.8-27B}}\\
    Syllogism prompting~\cite{jiang2023legal} & 56.25{\tiny$\pm$8.47} & 62.50 & 71.09 & 39.84 & 57.42 & 90.00 & 56.04 & 67.97{\tiny$\pm$7.98} & 17.97 & 14.06 & 32.03 & 0.122{\tiny$\pm$0.291} & 8.59 & 10.94 & 13.28 & 29.31 \\
    Syllogistic retrieval~\cite{zhang2025syler} & 67.19{\tiny$\pm$8.03} & 72.66 & 79.69 & 55.47 & 68.75 & 92.47 & 69.61 & 77.34{\tiny$\pm$7.19} & 14.06 & 8.59 & 22.66 & 0.241{\tiny$\pm$0.333} & 19.53 & 26.56 & 22.66 & 50.00 \\
    Agentic legal search~\cite{yang2026glare} & 67.97{\tiny$\pm$7.98} & 78.12 & 82.03 & 57.81 & 71.48 & 87.00 & 70.48 & 80.47{\tiny$\pm$6.83} & 17.97 & 1.56 & 19.53 & 0.840{\tiny$\pm$0.227} & 20.31 & 28.91 & 25.78 & 56.90 \\
    Element agents~\cite{yuan2024can} & 73.44{\tiny$\pm$7.57} & 80.47 & 84.38 & 68.75 & 76.76 & 91.26 & 81.48 & 82.03{\tiny$\pm$6.62} & 14.84 & 3.12 & 17.97 & 0.652{\tiny$\pm$0.302} & 25.78 & 39.84 & 27.34 & 60.34 \\
    Courtroom simulation~\cite{zhang2026chinese} & 64.06{\tiny$\pm$8.20} & 77.34 & 80.47 & 58.59 & 70.12 & 82.83 & 72.82 & 78.12{\tiny$\pm$7.10} & 16.41 & 5.47 & 21.88 & 0.500{\tiny$\pm$0.307} & 16.41 & 29.69 & 23.44 & 51.72 \\
    Multimodal agents~\cite{kang2026multimodal} & 74.22{\tiny$\pm$7.50} & 78.12 & 82.81 & 64.06 & 74.80 & 95.00 & 77.36 & 78.91{\tiny$\pm$7.01} & 13.28 & 7.81 & 21.09 & 0.259{\tiny$\pm$0.342} & 26.56 & 35.16 & 24.22 & 53.45 \\
    \midrule
    \textbf{\method{} (ours)} & 78.91{\tiny$\pm$7.01} & 83.59 & 90.62 & 70.31 & 80.86 & 94.39 & 77.59 & 85.16{\tiny$\pm$6.15} & 9.38 & 5.47 & 14.84 & 0.263{\tiny$\pm$0.398} & \textbf{31.25} & \textbf{41.41} & \textbf{30.47} & \textbf{67.24} \\
    \bottomrule
  \end{tabular*}
  \vspace{-2mm}
\end{table*}

\noindent\textbf{Inputs to the Judgment.}
The judge agent receives the closed graph and the legal materials, which fix what the case is and which statute governs it, and \emph{contextual skills} carrying what comparable circumstances have led courts to do (Sec.~\ref{ssec:skill}):
\begin{equation}
  \hat{J}=\argmax_{J\in\mathcal{J}(\mathcal{G}_\tau)}\ \pi_{\mathrm{judge}}\bigl(J\gv\mathcal{G}_\tau,\ X_L,\ \mathcal{S}_{\mathrm{ctx}}(z)\bigr),
  \label{eq:judge}
\end{equation}
where the context $z$, the triple (jurisdiction, offence family, seriousness band), is read off the closed graph.
Checking the elements is one step of this decision, not the whole of it: the same closed graph maps to different dispositions once aggravating and mitigating circumstances, seriousness and local practice are weighed~\cite{sentencingcouncil2019}.

\noindent\textbf{Syllogistic Form and Grounding.}
The harness restricts the judge to the syllogistic form, the cited statute as major premise and the established facts as minor, giving the feasible set:
\begin{equation}
  \begin{aligned}
    \mathcal{J}(\mathcal{G}_\tau)&=\bigl\{J:\ \mathrm{Facts}(r)\subseteq\{f_v\}_{v\in\mathcal{V}},\\
    &\qquad\ \forall\,\ell\in\mathrm{Elem}(c)\ \exists\,v\in\mathrm{Cites}(r):\ u_v=\ell\bigr\},
  \end{aligned}
  \label{eq:feasible}
\end{equation}
where $\mathrm{Facts}(r)$ are the factual claims in the reasoning, $\mathrm{Elem}(c)$ the elements the cited statute requires, and $\mathrm{Cites}(r)$ the nodes it cites.
The form holds the reasoning to what was established while $\mathcal{S}_{\mathrm{ctx}}$ supplies the weighing of circumstances, the two moves the poles of Fig.~\ref{fig:overview}a each leave out; infeasible judgments are rewritten.

\subsection{Evidence-Driven Skill Evolution}
\label{ssec:skill}

\noindent\textbf{Skills Distilled from Trajectories.}
The harness mines its own trajectories: recurring routes from evidence to a finding become fact-finding skills, recurring circumstance--outcome associations contextual skills, each stored with its context $z$~\cite{liao2026verdict,geng2026parthenon}.
Unlike libraries gated on task reward~\cite{wu2026bayesian}, every invocation $e$ of $u$ in $z$ is labelled by a verifier that checks that step alone and never sees the judgment's score, so a lucky outcome earns no credit:
\begin{equation}
  y_e\sim\mathrm{Bernoulli}(p_{u,z}),\qquad c_e\in[0,1],
  \label{eq:verifier}
\end{equation}
with $p_{u,z}$ the unknown reliability of $u$ in $z$ and $c_e$ the verifier confidence, low-confidence labels counting less in Eq.~\eqref{eq:beta}.

\noindent\textbf{Hierarchical Posterior.}
A two-level Beta prior lets sparse contexts borrow from the skill's other contexts and keeps the posterior closed-form:
\begin{equation}
  \begin{aligned}
    \mu_u&\sim\mathrm{Beta}\bigl(\kappa_0\mu_0,\ \kappa_0(1-\mu_0)\bigr),\\
    p_{u,z}&\sim\mathrm{Beta}\bigl(\kappa_u\mu_u,\ \kappa_u(1-\mu_u)\bigr),\\
    \alpha_{u,z}&\!=\!\kappa_u\mu_u\!+\!\textstyle\sum_{e}\!c_e y_e,\
    \beta_{u,z}\!=\!\kappa_u(1{-}\mu_u)\!+\!\textstyle\sum_{e}\!c_e(1{-}y_e),
  \end{aligned}
  \label{eq:beta}
\end{equation}
where $E_{u,z}$ are the verified invocations, $\mu_u$ the skill reliability, $\kappa_u$ the shrinkage and $(\mu_0,\kappa_0)$ the library prior; evidence is the effective sample size $n_{u,z}=(\sum_e c_e)^2/\sum_e c_e^2$.

\noindent\textbf{Credible-Bound Decisions.}
Decisions act on the credible interval rather than the posterior mean, its lower and upper bounds being Beta quantiles $Q_\delta$:
\begin{equation}
  \LCB_{u,z}=Q_{\delta}(\alpha_{u,z},\beta_{u,z}),\;
  \UCB_{u,z}=Q_{1-\delta}(\alpha_{u,z},\beta_{u,z}),
  \label{eq:bounds}
\end{equation}
and an operator $\Phi$ updates the library once per phase by the rule
\begin{equation}
  \Phi_u=\begin{cases}
    \mathrm{retain}, & \LCB_{u,z}\ge\theta,\\
    \mathrm{defer},  & \LCB_{u,z}<\theta\le\UCB_{u,z}\ \text{or}\ n_{u,z}<n_{\min},\\
    \mathrm{refine}, & \UCB_{u,z}<\theta,\ \omega_u\ge\omega_{\min},\\
    \mathrm{prune},  & \UCB_{u,z}<\theta,\ \omega_u<\omega_{\min},
  \end{cases}
  \label{eq:phi}
\end{equation}
with $\theta$ the reliability threshold, $n_{\min}$ the evidence floor and $\omega_u$ the usage share.
The interval tells insufficient evidence from confirmed failure: two failures defer a skill, seventeen in twenty refine it.
$\Phi$ also \textrm{splits} $u$ when $\Pr(p_{u,z}>p_{u,z'})\ge 1-\delta$, as when a rule valid in one jurisdiction lapses in another, and \textrm{generates} a skill where no $\LCB_{u,z}$ reaches $\theta$.
A candidate library is accepted only if its paired mean gain on held-out development cases is non-negative on Acc, Disp and Sent., else rolled back; scores gate the library, never a single skill.

\section{Experiments and Analysis}
\label{sec:exp}

All columns are defined in the caption of Table~\ref{tab:main}, and are read off one end-to-end judgment.
Open-weight and commercial multimodal LLMs (MLLMs) are evaluated under one protocol: the evaluation package stays closed, retrieval runs over the provided $X_L$ with no network lookup, and every model receives the same $\X$; audio-less backbones receive its written description.
\method{} plugs into the Qwen3.8-27B backbone with the same prompts; its library is distilled on a development split whose held-out part serves the acceptance test of Sec.~\ref{ssec:skill}, then frozen; the evaluation split enters no library decision.
Peer frameworks cover the prompting, retrieval and multi-agent families on the same backbone~\cite{jiang2023legal,zhang2025syler,yang2026glare,yuan2024can,zhang2026chinese,kang2026multimodal}.

\noindent\textbf{Where systems sit on the axis.}
The sign of Pol.\ separates the two failures: the three open-weight backbones are negative, arguing from grounds the record does not support, while every commercial model and every peer framework is positive, returning the statutory default instead.
Placement and charge accuracy rank-correlate at $0.98$ over the fourteen systems, so the axis tracks competence rather than replacing it.
Acc/Fam stays above $82\%$ everywhere, so a missed charge is usually the wrong offence in the right family; Sent/Disp instead divides by scale, under half for the open-weight backbones against $78$--$88\%$ for the commercial ones.

\noindent\textbf{\method{} performance.}
On its own backbone \method{} raises Acc from $47.7$ to $78.9$ and the $\Jgold$ share from $54.7\%$ to $85.2\%$, and turns Pol.\ from $-0.34$ to $+0.26$, removing the unsupported-grounds failure rather than trading it for the rigid one.

It leads every peer framework on Acc, Avg and the $\Jgold$ share, though at $128$ cases the margins sit inside the Wilson intervals; by Avg it ranks fourth overall, above Claude Sonnet 5, at $27$B.

\noindent\textbf{Ablation.}
The recording matters most: replacing the audio by its written description costs $24.0$ points of Avg, more than twice any other component (Table~\ref{tab:abl}).
The Element Graph and syllogistic reasoning follow, and the two failures separate as components are removed: without the graph the misses drift to unsupported grounds, without the contextual skills to the statutory default; a frozen library loses most of what the skills add.

\begin{figure}[!t]
  \centering
  \includegraphics[width=\columnwidth]{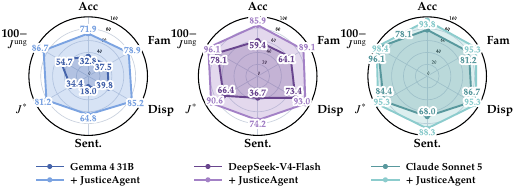}
  \vspace{-7pt}
  \caption{\method{} on three further backbones (128 cases, \%): backbone alone (dark), with \method{} (light).}
  \label{fig:radar}
\end{figure}

\begin{table}[!t]
  \centering\scriptsize
  \renewcommand{\arraystretch}{1.08}
  \vspace{-4pt}
  \caption{Ablation on the Qwen3.8-27B backbone (128 cases); columns as in Table~\ref{tab:main}, $\Delta$Avg the drop from the full harness. Each row drops the part of Sec.~\ref{sec:method} it names, the syllogistic form being Eq.~\eqref{eq:feasible}.}
  \label{tab:abl}
  \setlength{\tabcolsep}{0.9pt}
  \begin{tabular*}{\columnwidth}{@{\extracolsep{\fill}} l ccccc ccc c c}
    \toprule
    & \multicolumn{5}{c}{Correct (\%)} & \multicolumn{3}{c}{Nearest anchor (\%)} & & \\
    \cmidrule(lr){2-6}\cmidrule(lr){7-9}
    Variant & Acc & Fam & Disp & Sent. & Avg & $\Jgold$ & $\Jrig$ & $\Jung$ & Pol. & $\Delta$Avg \\
    \midrule
    Text-only input & 57.81 & 64.06 & 68.75 & 36.72 & 56.84 & 65.62 & 17.97 & 16.41 & 0.05 & -24.02 \\
    w/o skill evolution & 76.56 & 81.25 & 87.50 & 64.84 & 77.54 & 82.81 & 7.81 & 9.38 & -0.09 & -3.32 \\
    w/o syllogistic form & 72.66 & 78.12 & 82.03 & 54.69 & 71.88 & 78.91 & 10.16 & 10.94 & -0.04 & -8.98 \\
    w/o contextual skills & 74.22 & 79.69 & 86.72 & 64.84 & 76.37 & 82.03 & 10.94 & 7.03 & 0.22 & -4.49 \\
    w/o Element Graph & 71.09 & 75.00 & 81.25 & 55.47 & 70.70 & 78.12 & 8.59 & 13.28 & -0.21 & -10.16 \\
    \midrule
    \textbf{Full \method{}} & \textbf{78.91} & \textbf{83.59} & \textbf{90.62} & \textbf{70.31} & \textbf{80.86} & \textbf{85.16} & \textbf{9.38} & \textbf{5.47} & \textbf{0.26} & -- \\
    \bottomrule
  \end{tabular*}
  \vspace{0mm}
\end{table}

\noindent\textbf{Across backbones.}
The gain is largest where the backbone is weakest (Fig.~\ref{fig:radar}): Gemma 4 31B gains $39.1$ Acc, and Claude Sonnet 5 reaches $93.8$ Acc, above every stand-alone model of Table~\ref{tab:main}.
Sent.\ moves most on all three.

\section{Conclusion}
\label{sec:concl}
We formalized adjudication as a judgment anchored between rigid rule application and ungrounded discretion, scored against three lawyer-written references in \bench{}.
The failures divide by scale, and \method{} closes part of the gap with an Element Graph and verified experience.
The benchmark is retrospective and criminal only, and the harness is decision support, not a court.

\vfill\pagebreak
\section{Compliance with Ethical Standards}
This research study was conducted retrospectively using human subject data made available in open access by the courts and public agencies that released the recordings, footage and court records used here. Ethical approval was not required, as the study uses only publicly released material; participants are aliased in all annotations.

\bibliographystyle{IEEEbib}
\bibliography{refs}

\end{document}